\documentclass{article}
\usepackage[preprint]{neurips_2026}

\usepackage[utf8]{inputenc} 
\usepackage[T1]{fontenc}    
\usepackage{hyperref}       
\usepackage{url}            
\usepackage{booktabs}       
\usepackage{amsfonts}       
\usepackage{nicefrac}       
\usepackage{microtype}      
\usepackage{xcolor}         
\usepackage{amsmath}
\usepackage{amssymb}
\usepackage{tikz}
\usetikzlibrary{trees, shapes.geometric, arrows}
\usepackage{algorithm}
\usepackage{algorithmic}
\usepackage{makecell}
\usepackage{multirow}
\usepackage{subcaption}
\usepackage{rotating}
\usepackage{caption}
\title{Conditional Attribute Estimation with Autoregressive Sequence Models}

\author{
  Erica Stutz \\
  Department of Biomedical Informatics and Data Science\\
  Yale University\\
  New Haven, CT 06510 \\
  \texttt{erica.stutz@yale.edu} \\
\And
Giacomo Marino\\
  Department of Biomedical Informatics and Data Science\\
  Yale University\\
  New Haven, CT 06510 \\
  \texttt{giacomo.marino@yale.edu} \\
\And
Daniella Meeker\\
  Department of Biomedical Informatics and Data Science\\
  Yale University\\
  New Haven, CT 06510 \\
  \texttt{daniella.meeker@yale.edu} \\
\And
Qiao Liu\\
  Department of Biostatistics\\
  Yale University\\
  New Haven, CT 06510 \\
  \texttt{qiao.liu@yale.edu} \\
\And
Andrew J. Loza\\
  Department of Biomedical Informatics and Data Science, Department of Pediatrics\\
  Yale University\\
  New Haven, CT 06510 \\
  \texttt{andrew.loza@yale.edu} \\
}

\begin{document}

\maketitle

\begin{abstract}

Generative models are often trained with a next-token prediction objective, yet many downstream applications require the ability to estimate or control sequence-level properties. Despite their success, next-token prediction can lead to overfitting of local patterns during training, underfitting of global structure, and requires significant downstream modifications or expensive sampling to guide or predict the global attributes of generated samples at inference time. Here, we introduce Conditional Attribute Transformers, a novel method for jointly estimating the next-token probability and the value of an attribute conditional on each potential next token selection. This framework enables three critical capabilities within a single forward pass, without modification of the input sequence: (1) per-token credit assignment across an entire sequence, by identifying how each token in a sequence is associated with an attribute's value; (2) counterfactual analysis, by quantifying attribute differences conditional on alternative next token choices; (3) steerable generation, by decoding sequences based on a combination of next-token and attribute likelihoods. Our approach achieves state-of-the-art performance on sparse reward tasks, improves next-token prediction at sufficient model sizes, estimates attribute probabilities orders of magnitude faster than sampling, and can guide decoding of autoregressive sequence models on a range of language tasks.

\end{abstract}

\section{Introduction}

Generative models have demonstrated performance advances across multiple domains from language to biomedical informatics \cite{brown2020language,ferruz2022protgpt2,waxler2025generative,brixi2025genome}. While next-token prediction is a scalable training objective \cite{hoffmann2022trainingcomputeoptimallargelanguage}, it optimizes for local coherence and can lead to greedy overfitting of local patterns and suboptimal prediction of critical branch tokens \cite{gloeckle2024betterfasterlarge,qi2020prophetnet}. Furthermore, the downstream utility of these models is frequently defined by sequence-level attributes \cite{chang2023surveyevaluationlargelanguage}.  In language models, there is need to control token selection to create text with specific attributes, such as correctness or helpfulness \cite{keskar2019ctrl}. In biomedical informatics, generative models are used to estimate clinically relevant sequence attributes such as disease onset, medical events, or treatment response through expensive Monte Carlo (MC) simulation \cite{waxler2025generative,renc2024zero,shmatko2025learning}. The capacity to learn representations that better capture sequence-level properties, estimate properties efficiently from partial sequences, and control generation to create sequences with specific properties has broad implications across domains.

These diverse use cases share a common mathematical requirement: the need to estimate a sequence-level attribute for a partial sequence, conditional on the selection of the next token. This capability allows for estimation of attribute likelihood for observed sequences or steerable decoding to optimize the likelihood of a particular attribute. Current methods for predicting or controlling sequence-level attributes are computationally demanding because they often require properties that may not emerge until many tokens in the future. Current approaches fall into two main categories: conditioning and base-model steering with auxiliary models. These methods have several limitations including high computational overhead, requirements for additional model training, and limited flexibility.

\begin{figure} [htpb]
    \centering
    \includegraphics[width=1.0\linewidth]{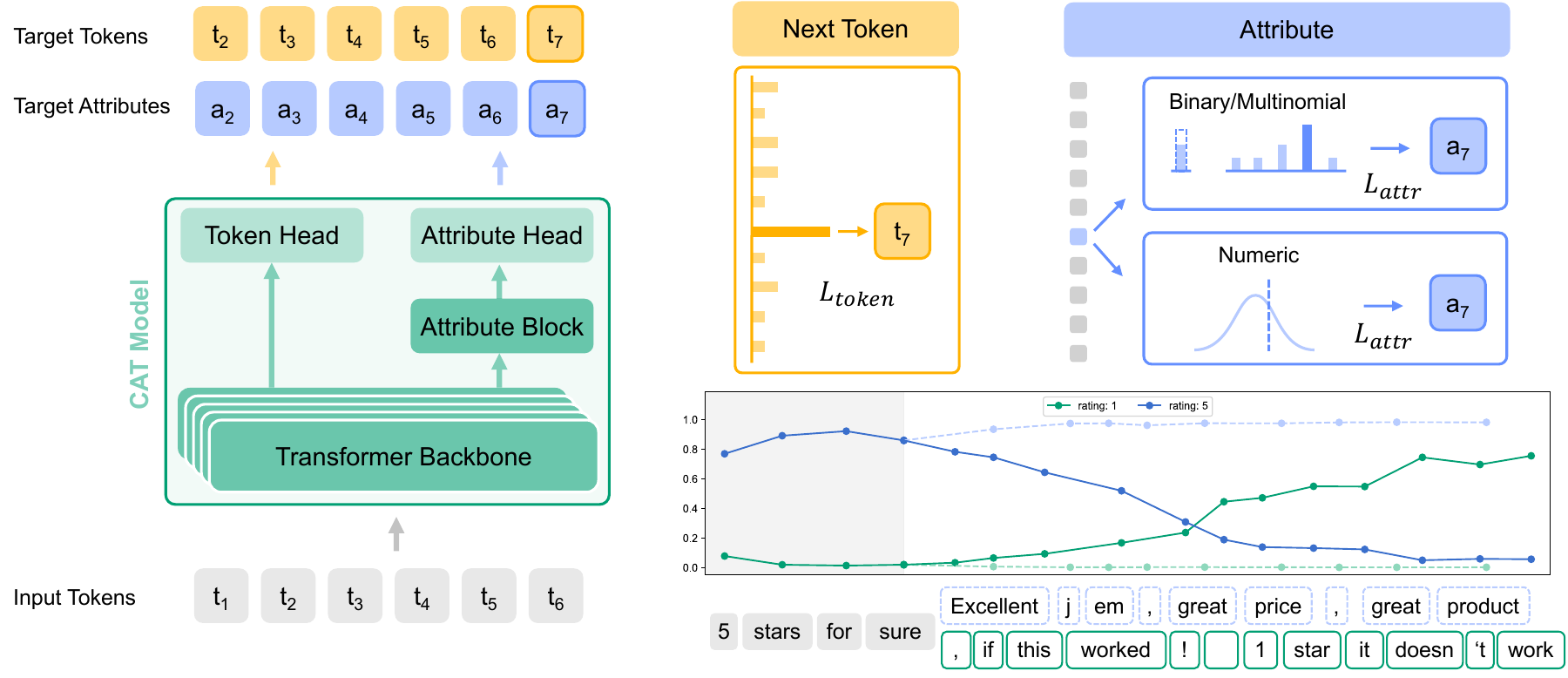}
    \caption{CAT is a unified architecture for next-token and sequence-level attribute prediction. Tokens ($t_n$) are processed by a shared backbone. The final latent representation is simultaneously processed by the language modeling head (Token Head) and an integrated conditional attribute model (attribute block + attribute head). Next-token cross-entropy loss ($L_{token}$) is combined with the attribute loss ($L_{attr}$) which can be from a binary, multinomial, or numeric attribute, delivered as a token-level sequence ($a_n$). During training, the full conditional attribute matrix does not have to be materialized (gray) because only the attribute of the true next token is seen, although it is learned. Lower right panel shows example of continuation (dashed) or CAT steering (solid) to change a 5 star prefix into a 1 star full review.} 
    \label{fig:diagram}
\end{figure}

Here, we propose Conditional Attribute Transformers (CAT), a method for conditional attribute estimation (Fig.~\ref{fig:diagram}). Leveraging a reinforcement learning framework, we cast data as arising from an unknown sequential game. We develop a generative modeling of this framework with a joint objective of next-token prediction and conditional sequence-level attribute prediction in a single model through a branched architecture and shared latent space.

The specific contributions of this work are as follows:
\begin{itemize}
    \item We provide a framework for simultaneous estimation of next-token likelihood and a sequence-level attribute likelihood, conditional on each potential next token. We also link this objective to components of causal inference and reinforcement learning.
    \item We demonstrate that this objective can be integrated into (a) the pre-training of generative decoder-only transformers with minimal computational overhead and can synergistically improve next-token perplexity, or (b) pre-trained models through fine-tuning.
    \item We evaluate performance on three diverse tasks: (1) learning strategy from random gameplay, (2) predicting and controlling the likely rating for Amazon product reviews, and (3) predicting sepsis onset in a medical data set.
\end{itemize}

\section{Related Work}
Numerous methods for controlling or estimating sequence-level attributes have been developed and fall into two classes: generation via conditioning or inference time guidance by a separate model.

Conditioning is used for generative model steering by inserting a fixed prompt or code in its input. Pre-training conditioning includes methods like CTRL \cite{keskar2019ctrl}, which prepends control codes to sequences, and Decision Transformers \cite{chen2021decision}, which insert reward-to-go tokens within each reward-state-action tuple using an offline reinforcement learning framework. Among post-training conditioning methods, Quark prepends sequences with reward quantile tokens \cite{lu2022quark}. While these models can condition generation, they do not ensure that insertion of control tokens produces sequences that remain in distribution \textit{or} provide probabilistic estimates of the attribute likelihood. Furthermore, if a downstream token is selected in error, there is no means of correction. CAT, by contrast, estimates the attribute at every step without requiring modification of the input sequence, allowing for more flexible and active control of steering.

Alternatively, token generation can be guided with auxiliary models. Classifier-based methods include PPLM, which uses gradients from an external classifier to guide generation, and FUDGE, which trains a binary classifier to predict attribute selection from partial sequences \cite{dathathri2019plug,yang2021fudge}. GeDi uses a generative discriminator to update next-token probabilities \cite{krause2020gedi}. Director integrates the generator and classifier into a unified model by adding an attribute head alongside the language modeling head at the final latent representation \cite{arora2022director}. DExperts uses two auxiliary models, one fine-tuned on the desired attribute and the other on an undesired attribute, to reweight next-token probabilities of the main model \cite{liu2021dexperts}.  TRACE distills a hidden Markov model from a base language model to calculate the sequence-level attribute probability, while ILQL takes this a step further by using full Q-learning and not just single-step policy updates \cite{weng2025trace, snell2022offline}. Many of these approaches require significant computational overhead due to the training of auxiliary models including three separate transformer models with ILQL \cite{snell2022offline}. Notable exceptions are TRACE and Director; however, they are limited by the expressivity of the distilled hidden Markov model and a simple linear layer. CAT maintains the full expressivity of a transformer model, yet requires substantially less computational overhead than training a full auxiliary model.

\section{Methods}
\label{sec:model}
\subsection{Definitions}

We model data as arising from an unknown sequential game in which the rules, players, strategies, and differences between actions and observations are unknown. We only have access to records of its play, which include a mixed sequence of actions and observations and an associated result which we consider as the sequence-level attribute, $\alpha$. The attribute can be binary, multinomial, or numeric. After observing these records, we aim to address two questions: (1) can an agent learn to play (i.e. to generate valid moves) and (2) can it learn to play well (i.e. to achieve a specified result)?

Each game consists of a sequence $S = [s_1, s_2,..., s_N]$ of observations or actions $s_i$. We do not assume that we need to differentiate between these two during training. We use $s_n$ to denote a candidate next element in the sequence. For simplicity, we will consider only $s_n$ that are drawn from a discrete language $L$. For a sequence $S$, there exists an unknown function $f(S) = P(\alpha_i \mid S) $ that generates $\alpha_i \in A$ conditioned on $S$, which we take here to be the result of the game. We assume that each sequence $S$ represents a valid game instance, but make no assumption that it is generated from a specific policy. Instead, it can be viewed as a sample from an average policy distribution $\pi^{\mu}$.

\subsection{Model}
\paragraph{Joint Distribution:} 
A record of an unknown game can be drawn from the joint distribution of the sequence of gameplay and its result attribute, $P(S,\alpha_i)$. A standard autoregressive decomposition is:
\begin{align}
    P(S,\alpha_i) &= P(\alpha_i \mid S)P(S)\\
    P(S) &= \prod_{i=1}^k P(s_i \mid s_1, ..., s_{i-1})
\end{align}
If $\alpha_i$ is treated as an additional observation, a single autoregressive model can be used for actions, game state, and game result. However, the probability of $\alpha_i$ can only be estimated after conditioning on the full sequence. For partial sequences, MC simulation must be used to estimate $P(\alpha_i, S)$.

\paragraph{Alternative Decomposition:}
We can consider $S$ to be a sequence of three parts $(S_a,s_n,S_b)$, a prefix sequence $S_a$ of all observations preceding $s_n$, a next observation $s_n$, and a suffix sequence $S_b$ consisting of all observations following $s_n$. The original joint distribution can be expanded as:
\begin{equation}
P(\alpha_i, S) = P(\alpha_i, S_a, s_n, S_b)
\end{equation}
We can decompose this new expanded joint distribution as:
\begin{equation}
P(\alpha_i, S_a, s_n, S_b) = P(S_a) \cdot P(s_n \mid S_a) \cdot P(\alpha_i, S_b \mid S_a, s_n)
\end{equation}
We can then marginalize over $S_b$:
\begin{equation}
\sum_{S_b} P(\alpha_i, S_a, s_n, S_b) = P(S_a) \cdot P(s_n \mid S_a) \cdot \sum_{S_b} P(\alpha_i, S_b \mid S_a, s_n)
\end{equation}
which yields:
\begin{equation}
P(\alpha_i, S_a, s_n)
= \underbrace{P(S_a)}_{\mathrm{prefix}}
\cdot
\underbrace{P(s_n \mid S_a)}_{\mathrm{sequence\ model}}
\cdot
\underbrace{P(\alpha_i \mid S_a, s_n)}_{\mathrm{attribute\ model}}
\end{equation}
Since we make no assumptions on the length of $S_b$, this decomposition is valid for any position of $s_n$ including the end of the sequence $S$, in which case $S_b = \varnothing$. 

\paragraph{Distribution Estimation:} This decomposition can be modeled using an augmented causal transformer with two heads: a next-token prediction head for estimating $P(s_n \mid S_a)$ and conditional attribute prediction head for estimating $P(\alpha_i \mid S_a,s_n)$. The prefix $P(S_a)$ is modeled autoregressively by the sequence head, enabling sampling from the full joint distribution $P(\alpha_i, S)$. Here, we use a shared model backbone $f_{\theta}$ to produce a hidden representation $H$. 
\begin{equation}
f_{\theta}(S) = H, \quad \text{where } f_{\theta}: S \to \mathbb{R}^d
\end{equation}
This representation is passed to two heads:
\begin{equation}
g_{\psi}(H) = P(s_n \mid S), \quad \text{where } g_{\psi}: \mathbb{R}^d \to \Delta^{|L|}
\end{equation}
\begin{equation} 
h_{\phi}(H, s_n) = P(\alpha_i \mid S, s_n), \quad \text{where } h_{\phi}: \mathbb{R}^d \times L \to 
\begin{cases}
\Delta^{|A|} & \text{(categorical attributes)} \\
\mathbb{R}^p & \text{(parameterized attributes)}
\end{cases}
\end{equation}
such that the backbone model $f_\theta$ contains information about both next-token and attribute while $g_\psi$ and $h_\phi$ provide task-specific transformations. The function $h_\phi$ can be selected to produce attribute category logits or distributional parameters depending on the attribute type. Here, we demonstrate the model using binary, multinomial, and numeric attributes.

\subsection{Inference}
Samples of observed games can be generated via $s_n \sim P(s_n \mid S_a)$ from the sequence model. The conditional attribute model can be used for alternative sampling strategies such as greedy decoding toward an attribute of interest. When selecting $s_n$, we can restrict choices to estimated valid moves (next-token probability above a threshold $\epsilon$) and select one that maximizes the probability of achieving a desired result while remaining within the distribution of the training data to a certain threshold $\epsilon$:
\begin{equation}\label{optimalsn}
    s_n^* = \arg \max_{s_n} P(\alpha_i \mid S_a,s_n)  \quad \text{for all} \quad s_n \in P(s_n \mid S_a) > \epsilon
\end{equation}
Eq.~\ref{optimalsn} defines a greedy optimizer for maximizing the likelihood of $\alpha_i$. This approach can be used to select locally optimal actions and develop a policy $\pi^{new}$ that improves upon the average policy $\pi^{\mu}$. However, it does not guarantee attainment of the global optimum.

\subsection{Relationship to Other Entities}
While we introduce this model as a conditional attribute estimator, it is directly related to functions in reinforcement learning and causal inference.

\paragraph{Relationship to Q-Functions:}
Although derived using different notation, the components of this framework map to those in Distributional Reinforcement Learning in an offline setting \cite{bellemare2017distributionalperspectivereinforcementlearning}. Consider the case where $\alpha_i$ represents a reward, $s_n$ represents an action, and $S_a$ represents a state. Note that here we have discussed the attribute as a terminal reward, but the derivation applies to sub-sequences as well. The sequence model $P(s_n \mid S_a)$ functions as the average behavior policy $\pi_\mu(s_n \mid S_a)$, representing the probability of selecting an action $s_n$ given a state $S_a$ in the training distribution. The attribute model $P(\alpha_i \mid S_a, s_n)$ learns the conditional distribution over rewards. The standard state-action value function under the behavior policy, $Q^{\pi_\mu}(s, a)$, is recovered by taking the expectation over this distribution:
\begin{equation}
Q^{\pi_\mu}(S_a, s_n) = \mathbb{E}_{\alpha \sim P(\cdot \mid S_a, s_n)}[\alpha]
\end{equation}
We note that this work focuses on single-step policy improvement by selecting from the learned $Q$ function in a greedy (or adjusted-greedy manner) rather than learning a globally optimal $Q^*$ through recursive value updates \cite{snell2022offline}.

\paragraph{Relationship to Causal Models:}
In the case where $s_n$ represents a treatment or intervention and $S_a$ represents the history of confounders or covariates, this approach parallels the core components of causal inference. Specifically, the sequence model $P(s_n \mid S_a)$ functions as a generalized propensity score, estimating the probability of treatment assignment conditioned on covariates:
\begin{equation}
e(S_a) = P(Z = s_n \mid X = S_a)
\end{equation}
Simultaneously, the attribute model $P(\alpha_i \mid S_a, s_n)$ functions as a conditional outcome model (or response surface), estimating the expected outcome given the covariates and treatment:
\begin{equation}
\mu(S_a, s_n) = \mathbb{E}[Y = \alpha_i \mid X = S_a, Z = s_n]
\end{equation}
While it operates on fixed covariates, focuses on single treatment assignment, and uses binary outcomes, Dragonnet is similar in that it uses a shared latent representation for prediction of a propensity and outcome model \cite{shi2019adapting}. The recent generative causal inference approaches \cite{liu2024encoding, liu2026ai,liu2026bayesian}, further generalize this idea through latent probabilistic modeling of treatments, covariates, and outcomes, enabling flexible conditional inference and uncertainty quantification. Causal transformers use sequential data, but require separate streams for covariates, treatments, and outcomes recorded at specific time intervals \cite{melnychuk2022causal}. Other methods use expensive MC rollouts for estimating the outcomes associated with treatment assignments \cite{rein2024deep,xiong2024g}. CAT improves upon these by using full sequential data and avoiding MC simulation.

\subsection{Model Architecture}
We extended the nanoGPT architecture \cite{karpathy2023nanogpt}, for joint next-token and conditional attribute prediction. The forward pass of CAT matches  standard transformers until the final latent embedding. This branches to a standard language modeling head and the conditional attribute transformer block (similar to how multi-token prediction is performed with DeepSeek-R1 during pretraining) \cite{liu2024deepseek}. This is used to predict the parameters required for attribute prediction which can be binary, multinomial, or numeric. Computationally efficient training is achieved by recognizing that although next-token requires computing logits across the entire vocabulary, the conditional attribute loss only requires computing the probabilities of attributes for the true single next token. Therefore, for a vocabulary of size $V$ and an attribute dimension of $A$, only a $1 \times A$ matrix must be materialized during training instead of a $V \times A$ matrix. Cross-entropy loss is used for next-token prediction and the attribute loss can be any likelihood-based loss. We used cross-entropy loss for binary and multinomial tasks, and Guassian negative log-likelihood for regression tasks. The total training loss was defined as:
\begin{equation}\label{lossfn}
    L = L_{token}+\lambda*L_{attr}
\end{equation}
where $L_{token}$ is next-token cross-entropy loss, $L_{attr}$ is attribute loss, and $\lambda$ controls the relative contribution of the attribute loss. Optimal $\lambda$ values varied across tasks, and were selected by balancing next-token loss, attribute loss, and sampling performance. Model sizes and hyperparameters for each experiment are provided in Table~\ref{app:key_door_model_architecture}, \ref{app:language_model_architecture}, and \ref{app:physionet_model_architecture}.

\subsection{Data Sets}
Experiments were conducted on three data sets: (1) the Key-to-Door environment used in the Decision Transformers paper \cite{chen2021decision} to test performance in a sparse-reward setting, (2) a language modeling data set (Amazon Reviews \cite{hou2024bridging}) to test performance on large action spaces, and (3) a clinical data set (PhysioNet Sepsis \cite{reyna2019early}) to test performance on a biomedical informatics task. 

\begin{enumerate}
    \item \textbf{Key-to-Door:} data on 10,000 random-walk trajectories in a three-room grid world consisting of a key room, a distractor room, and a door room. The agent must pick up the key and reach the door within a fixed move budget. This environment was introduced in \cite{mesnard2021counterfactualcreditassignmentmodelfree} and later used in \cite{chen2021decision} as a credit assignment benchmark.

    \item \textbf{Amazon Reviews:} data on 574 million Amazon product reviews from \cite{hou2024bridging}. Each sequence consists of a product category, title, review text, and a 1-5 star rating, which serves as a multinomial attribute.
    
    \item \textbf{PhysioNet Sepsis:} ICU time-series data set with 40,336 patient sequences for early sepsis identification. Each sequence contains patient demographics followed by hourly vital-sign and laboratory-measurement tokens, with time tokens separating hourly measurements. This was derived from the 2019 PhysioNet Challenge data set \cite{reyna2019early}.
\end{enumerate}

\section{Results}
\label{sec:results}

\subsection{Key-to-Door: Long-Term Credit Assignment}
\label{sec:key-to-door}
\begin{figure}[htbp]
    \centering
    \includegraphics[width=0.8\linewidth]{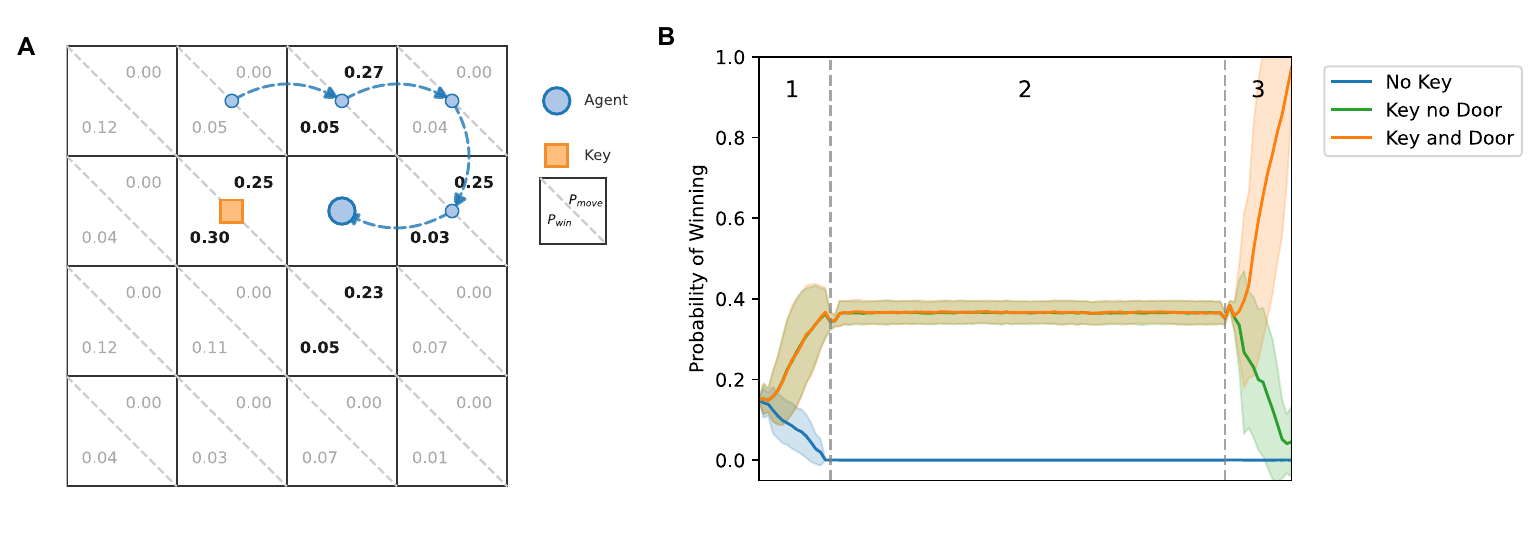}
    \caption{Key-to-Door task. \textbf{A} Agent moving in the key room with the move and win probabilities. \textbf{B} Average and 95\% confidence interval for estimated win probability for trajectories stratified by outcome. The dashed lined demarcates moves in each rooms: key (1), distractor (2), and door (3).}
    \label{fig:keytodoor}
\end{figure}

The Key-to-Door task tests CAT's ability to learn attribute assignment from a single terminal reward. CAT successfully learns to estimate conditional win probabilities for each potential next move (Fig.~\ref{fig:keytodoor}A) and can serve as a critic to stably classify the win probability for evolving trajectories with reduced variance compared to decision transformers (Fig.~\ref{fig:keytodoor}B)\cite{chen2021decision}. Using greedy decoding (Eq.~\ref{optimalsn}), the model outperforms all baselines: random policy, behavior cloning, percentile behavior cloning (trained on only winning examples), conservative Q-learning, and decision transformers (Table~\ref{tab:keytodoorwinrates}).

\begin{table}[htbp]
\caption{Win rates of evaluated methods for the Key-to-Door task. In addition to the 0.999 win rate, CAT takes the shortest manhattan-distance path 998 out of the 999 wins.}
\label{tab:keytodoorwinrates}
\centering
\scriptsize
\begin{tabular}{lcccccc}
\hline
\\[-0.8em]
\textbf{} 
& \shortstack{\textbf{Random}\\\textbf{Policy}}
& \shortstack{\textbf{Behavioral}\\\textbf{Cloning}}
& \shortstack{\textbf{Percentile}\\\textbf{Behavioral Cloning}}
& \shortstack{\textbf{Conservative}\\\textbf{Q-Learning}}
& \shortstack{\textbf{Decision}\\\textbf{Transformers}}
& \shortstack{\textbf{Conditional Attribute}\\\textbf{Transformers}} \\
\\[-0.8em]
\hline
\\[-0.8em]
\textbf{Win Rate} 
& 0.031 
& 0.016 
& 0.951 
& 0.133 
& 0.946 
& \textbf{0.999}
\\
\hline
\end{tabular}
\end{table}
\subsection{Language Modeling: Amazon Reviews}
\label{sec:amazon}
To test scalability and performance with a large input and action space, we evaluate CAT on the Amazon Reviews dataset, using review text as the sequence and the multinomial 5-class product rating as the attribute (Fig.~\ref{app:reviewoverview}). We evaluate (1) the impact of joint training on next-token prediction performance, (2) the performance of its role as a critic by estimating the rating from partial reviews, (3) the identification of semantic shifts caused by counterfactual adjective substitution, and (4) steering generation to create reviews with specific ratings. Full model details are reported in Appendix~\ref{app:amazon-reviews}.

\subsubsection{Next-Token Prediction Performance}
\begin{figure}[htbp]
    \centering
    \includegraphics[width=1.0\linewidth]{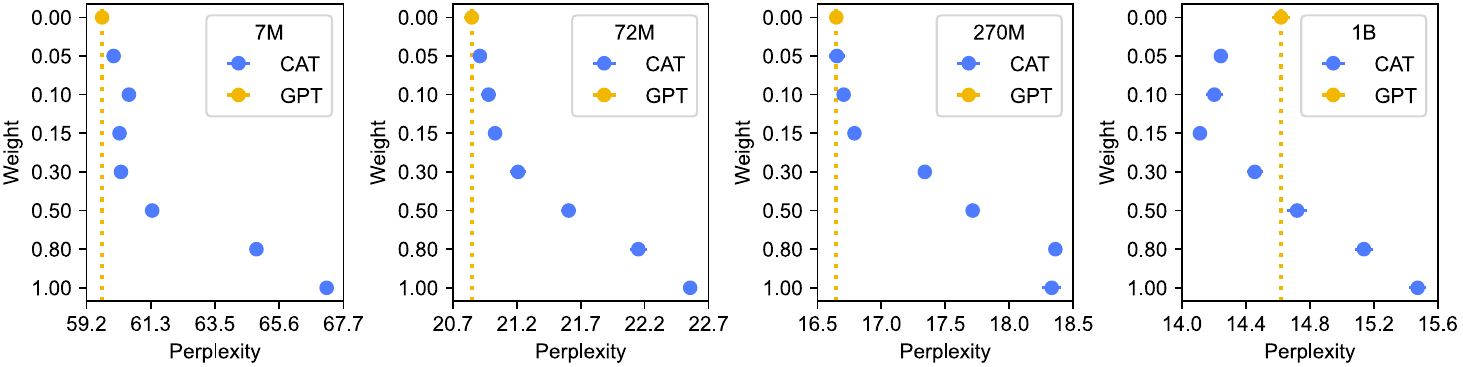}
    \caption{Token perplexity for CAT versus GPT. The perplexity from next-token prediction for CAT models versus standard GPT models across model size and CAT attribute head weights is shown. The GPT model is equivalent to a weight of zero (no attribute loss contribution). Synergy between the two tasks of next-token and conditional attribute prediction depends on model size and task weight.}
    \label{fig:amazonperplexity}
\end{figure}

In contrast to prior methods where joint-training worsened next-token prediction performance \cite{arora2022director}, we find that the CAT architecture enables improved next-token prediction performance with increasing model size. At small model sizes (7-72-million parameters), the conditional attribute task impairs next-token prediction. However, the 1-billion parameter CAT model achieved better performance on next-token prediction than the standard model. This synergy depended on $\lambda$ used to balance the two contributions to the joint loss (Fig.~\ref{fig:amazonperplexity}). Subsequent experiments use $\lambda=0.15$ to balance next-token and attribute prediction performance.
\subsubsection{Review Critic Performance}
\begin{figure}[htpb]
    \centering
    \includegraphics[width=0.7\linewidth]{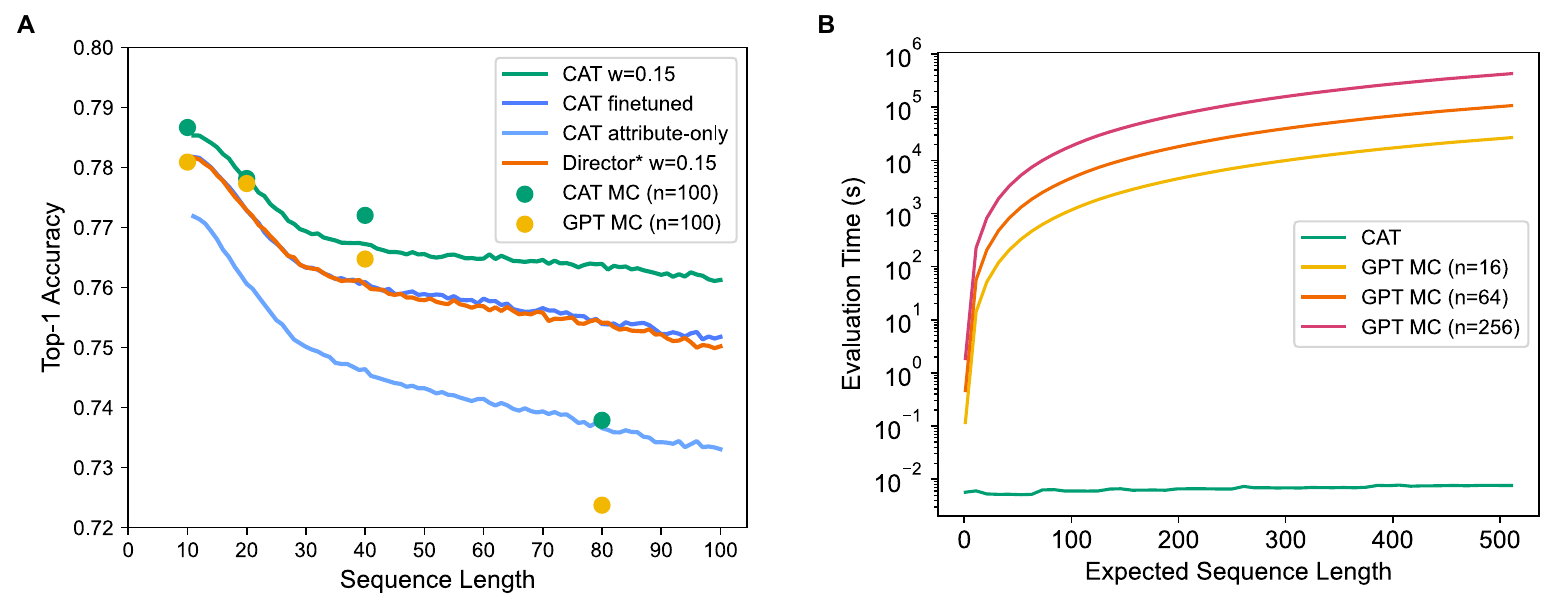}
    \caption{Rating prediction from partial reviews. \textbf{A} Top-1 rating prediction accuracy for CAT, fine-tuned CAT, attribute-only CAT, Director*, standard next-token MC simulation ($n=100$), and CAT MC simulation ($n=100$). The first four models were evaluated on 1 million reviews; sampling-based approaches were evaluated on 4,000 reviews due to computational cost. \textbf{B} Compute time as a function of expected sequence length for CAT and standard next-token MC simulation.}
    \label{fig:reviewcritic}
\end{figure}

CAT estimates attributes from partial sequences in a simulation-free manner. Performance of predicting the rating using CAT (conditional attribute probability for the true next token) was compared to MC simulation, an attribute head fine-tuned on a frozen standard GPT model, an attribute-only CAT model, and a version of Director which we extended for multinomial attributes (Director*) (Fig.~\ref{fig:reviewcritic}A). Due to the extensive compute required, MC sampling was evaluated on 4,000 reviews and four partial review lengths, whereas other methods are evaluated on 1 million reviews.

CAT and the fine-tuned CAT model both outperformed Director*, and MC sampling using CAT’s next-token head outperformed the standard next-token model. The attribute-only CAT model underperformed most methods, except standard next-token MC simulation at a partial sequence length of 80, indicating that jointly modeling next-token and conditional attribute prediction improves performance over either task alone. CAT also provided an approximately $10^8 \times$ speedup in partial sequence evaluation relative to MC sampling (Fig. \ref{fig:reviewcritic}B).

\subsubsection{Counterfactual Estimation}

\begin{table}[htbp]
\caption{Counterfactual estimation of substituting \textit{good} with alternative adjectives on predicted 1 and 5 star probabilities ($\Delta P_1$, $\Delta P_5$). Shown for all contexts and negated contexts (i.e. \textit{not} good)}.
\label{tab:counterfactual_estimation}
\centering
\scriptsize

\begin{tabular}{@{}lcccc@{}}
\toprule

\multirow{2}{*}{\textbf{Counterfactual}} 
& \multicolumn{2}{c}{\boldmath$\Delta P_1$}
& \multicolumn{2}{c}{\boldmath$\Delta P_5$} \\

\cmidrule(lr){2-3}
\cmidrule(lr){4-5}

& \shortstack{All\\Contexts} 
& \shortstack{Negation\\in Context} 
& \shortstack{All\\Contexts} 
& \shortstack{Negation\\in Context} \\

\midrule

good $\rightarrow$ AMAZING  & -0.01 & -0.07 &  0.09 & 0.02 \\
good $\rightarrow$ amazing  & -0.01 & -0.11 &  0.07 & 0.02 \\
good $\rightarrow$ GREAT    &  0.00 & -0.08 &  0.06 & 0.00 \\
good $\rightarrow$ great    &  0.00 & -0.08 &  0.04 & 0.00 \\
good $\rightarrow$ GOOD     &  0.00 &  0.00 &  0.03 & 0.00 \\

\midrule

good $\rightarrow$ bad       & 0.06 & -0.14 & -0.08 & 0.06 \\
good $\rightarrow$ BAD       & 0.08 & -0.11 & -0.10 & 0.03 \\
good $\rightarrow$ horrible  & 0.09 & -0.11 & -0.11 & 0.02 \\
good $\rightarrow$ HORRIBLE  & 0.10 & -0.04 & -0.19 & 0.01 \\

\bottomrule
\end{tabular}

\end{table}

We evaluated the semantic accuracy of CAT’s counterfactual estimates by substituting sentiment-bearing adjectives in 1,000,000 validation reviews. Replacing \textit{good} with negative adjectives increased 1 star and decreased 5 star probabilities. Positive substitutions increased 5 star probabilities with little effect on 1 star probabilities. Substitutions in negated contexts (\textit{not good}) had more nuanced shifts: both positive and negative substitutions reduced 1 star probability and led to no change or increased 5 star probabilities. Capitalization modulated these effects, reflecting the semantic emphasis (Table~\ref{tab:counterfactual_estimation}).

\subsubsection{Guided Decoding}

In steering 3 star prompts to 1 and 5 star reviews, CAT had the highest accuracy (CAT accuracy 0.64/0.77; best alternative Director* 0.58/0.65) across all evaluated models. Furthermore, generations were more fluent than models of similar accuracy (CAT perplexity 45.88/44.03; best alternative Director* 46.77/48.16) and had comparable diversity (Table~\ref{tab:steering}).

\begin{table}[htbp]
\caption{Steering 3 star prompts toward 1 and 5 star reviews.}
\label{tab:steering}
\centering
\scriptsize
\renewcommand{\arraystretch}{1.15}

\begin{tabular}{llcccccc}
\toprule

& \multirow{2}{*}{\textbf{Model}} 
& \multirow{2}{*}{\textbf{Accuracy}} 
& \multirow{1}{*}{\textbf{Fluency ($\downarrow$)}}
& \multicolumn{3}{c}{\textbf{Diversity ($\uparrow$)}} \\

& & 
& Mean ppl.
& Dist-1 
& Dist-2 
& Dist-3 \\

\midrule

\multirow{5}{*}{1 star}
& CTRL          & 0.13 & 13.68 & 0.90 & 0.87 & 0.78 \\
& DExperts      & 0.14 & 19.24 & 0.91 & 0.86 & 0.77 \\
& Director      & 0.25 & 20.18 & 0.88 & 0.89 & 0.82 \\
& Director*     & 0.58 & 46.77 & 0.80 & 0.80 & 0.75 \\
& \textbf{CAT}  & \textbf{0.64} & 45.88 & 0.73 & 0.65 & 0.57 \\

\midrule

\multirow{5}{*}{5 star}
& CTRL          & 0.41 & 13.45 & 0.90 & 0.86 & 0.77 \\
& DExperts      & 0.41 & 18.99 & 0.91 & 0.85 & 0.76 \\
& Director      & 0.47 & 18.95 & 0.90 & 0.86 & 0.78 \\
& Director*     & 0.65 & 48.16 & 0.83 & 0.75 & 0.65 \\
& \textbf{CAT}  & \textbf{0.77} & 44.03 & 0.79 & 0.84 & 0.82 \\

\bottomrule
\end{tabular}
\end{table}

\subsection{PhysioNet Sepsis}
\label{sec:sepsis}
\begin{figure}[htbp]
    \centering
    \includegraphics[width=0.8\linewidth]{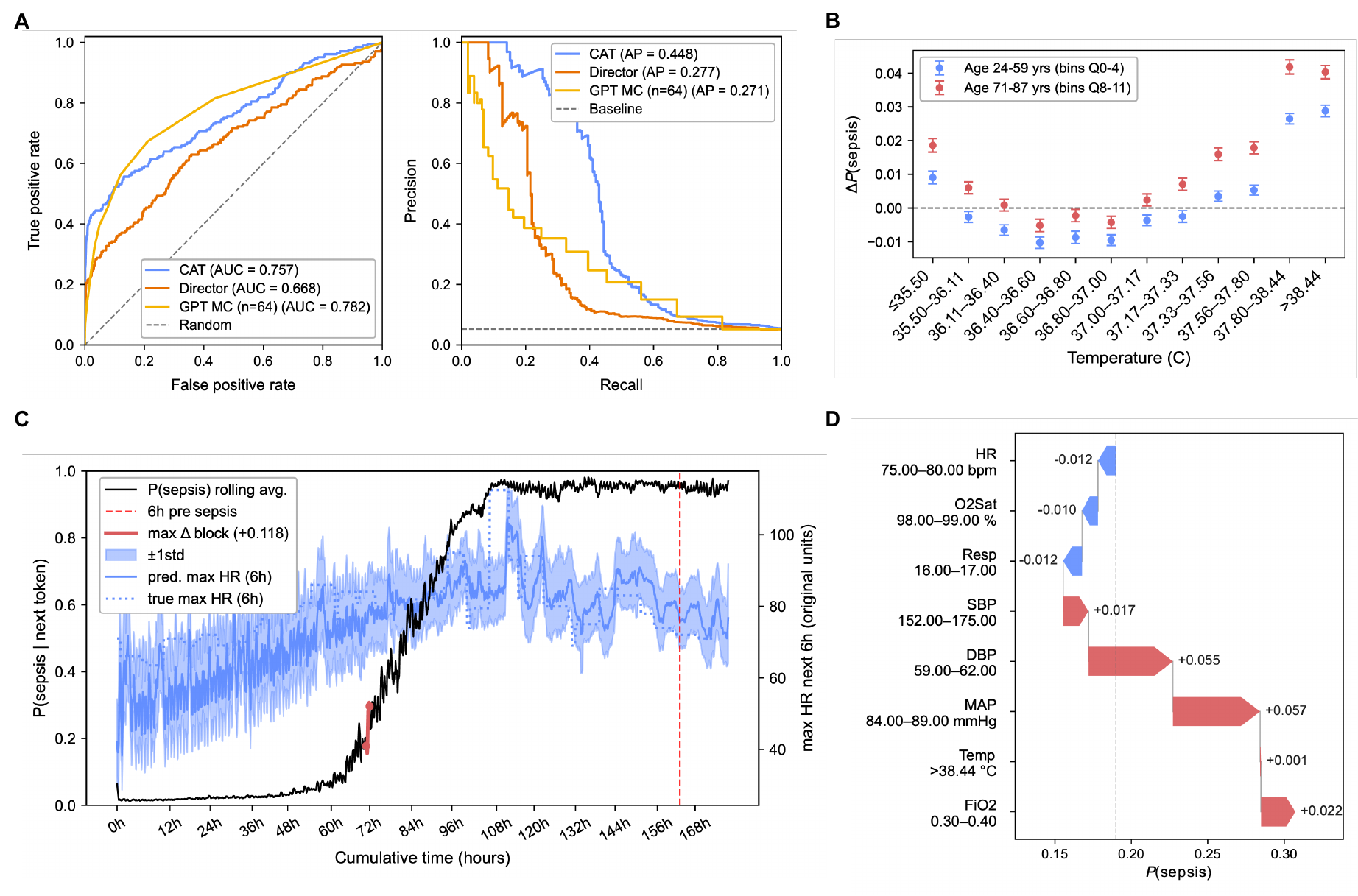}
    \caption{Predicting sepsis and maximum heart rate (HR) per token. \textbf{A} Receiver operating characteristic (ROC) and precision–recall (PR) curves on the PhysioNet 2019 sepsis task for CAT, Director, and GPT MC simulation (n=64). \textbf{B} Mean change in sepsis probability, $\Delta P(\mathrm{sepsis})$, as a function of the counterfactual vitals bin (mean $\pm$ SEM) for first temperature measurement in sepsis-positive sequence in validation data set, stratified by age (24–59 yrs vs 71–87 yrs). \textbf{C} Example validation case showing predicted sepsis probability (black) and predicted maximum HR in the next 6h with the maximum HR plotted with a dashed line (light blue). The red dashed line marks the 6-hours to sepsis time point. Both reported sepsis risk and predicted max HR are plotted as rolling averages with window of 5. \textbf{D} Per-token attribution for the largest 1-hour increase in sepsis risk, marked in red in \textbf{C}.}
    \label{fig:sepsis}
\end{figure}

To evaluate CAT’s critic performance beyond reinforcement learning and language, we assessed its ability to predict sepsis in ICU patients. We evaluated two sequence-level attributes: a binary label for sepsis occurrence during the ICU course, and a continuous label estimating maximum heart rate (HR) within a 6-hour sliding window. Assessed 12 hours prior to onset, CAT demonstrated comparable predictive performance compared to MC simulation of the standard next-token model and superior to Director (Fig.~\ref{fig:sepsis}A). Although the standard model achieved a slightly higher ROC AUC, CAT provides a substantial improvement in Average Precision (AP), indicating higher precision at clinically relevant recall levels in this highly imbalanced prediction setting.

To evaluate sensitivity to counterfactual vital sign substitutions, we isolated the first temperature reading in sepsis-positive validation sequences and measured stepwise changes in predicted sepsis probability across temperature bins (Fig.~\ref{fig:sepsis}B). Elevated temperature increased predicted sepsis risk, with substantially larger hyperthermic risk increases among older patients, consistent with their greater vulnerability to thermal dysregulation \cite{brody1994hyperthermia}. A representative case illustrates CAT’s dynamic sepsis risk estimates alongside conditional forecasts of maximum HR over the subsequent six hours (Fig.~\ref{fig:sepsis}C). Risk is assessed at each token, demonstrating an implicitly learned sensitivity to established clinical criteria. Extracting the hour with the largest increase in $P(\mathrm{sepsis})$ reveals the token-level contributions driving this risk change (Fig.~\ref{fig:sepsis}D). Large MAP-associated risk increases following low DBP may reflect discordance between MAP, SBP, and DBP measurements in the data set (Fig.~\ref{fig:sepsissupplement}).

\section{Discussion}
\label{sec:discussion}
We present Conditional Attribute Transformers (CAT), a novel framework to jointly model next-token probabilities and sequence level attributes for credit assignment, counterfactual estimation, and guided decoding. We find that with large enough models, CAT improves next token prediction, suggesting that forcing the model to learn global structure can complement rather than compromise next-token prediction. This mirrors results from the multi-token prediction training objective utilized by DeepSeek \cite{liu2024deepseek}, where forecasting future states acts as an auxiliary training objective, regularizing representations and encouraging broader structural planning. In addition, CAT's capabilities can be extended to pretrained models through fine-tuning only the conditional attribute head.

Across a range of settings and tasks, CAT consistently outperformed baseline models. In the Key-to-Door task, CAT learns long-term credit assignment from sparse terminal rewards and achieves near-perfect performance, demonstrating its ability to identify actions that determine delayed outcomes. In language modeling, CAT effectively captures shifts in review sentiment, enables efficient partial-sequence rating prediction, and supports attribute-guided generation. In biomedical informatics, CAT estimates elevated sepsis risk well before diagnosis is recorded, supports identification and examination of the drivers of major inflection points in predicted risk, and characterizes subgroup-specific changes in risk associated with clinical variables. While CAT offers profound societal benefits, particularly by advancing explainability through counterfactual analysis, it also carries inherent risks, as its steering capabilities could be exploited to introduce harmful or malicious biases.

\textbf{Limitations:} This method is currently limited to generating conditional attribute probabilities over discrete action choices, but there may be settings where a continuous attribute is desired. For steering problems, current counterfactual estimates represent single-step policy improvement and yield a greedy optimum under an average behavior policy rather than a global optimum under a specific policy. This limitation is shared by prior approaches such as Director and DExperts.

\textbf{Future Work:} Future work includes scaling this method to larger data sets and extending it to continuous action spaces. We are currently developing approaches to enable global optimal choice selection. Furthermore, this framework could be naturally extended to any task which requires estimation or control of sequence-level properties, including the biological domain (\textit{de-novo} protein design, predicting small-molecule binding functionality, predicting sequence-to-function regulatory mechanism from DNA, etc.).

Ultimately, by jointly modeling local token generation and global sequence level objectives, CAT provides a highly adaptable foundation for complex predictive and generative tasks.

\section{Acknowledgements}

This work was supported by the United States National Library of Medicine (grant T15 LM007056 to ES and GM) and National Institutes of Health (grants 5UL1TR001863 to DM and R00HG013661 to QL). AJL receives funding through UL1 TR001863, The Hartwell Foundation, and the ARIA foundation.

\bibliographystyle{unsrt}
\bibliography{references}

\appendix
\renewcommand{\thetable}{\Alph{section}.\arabic{table}}
\setcounter{table}{0}
\renewcommand{\thefigure}{\Alph{section}.\arabic{figure}}
\setcounter{figure}{0}

\section{Appendix}
\label{sec:appendix}

All experiments were run on a computing cluster with a combination of NVIDIA H100 and H200 and RTX5000 for approximately 3,000 GPU hours. Training and inference runtime varies with model architecture, model size, input sequence length, and hardware configuration (Table~\ref{app:key_door_model_architecture}, \ref{app:language_model_architecture}, and \ref{app:physionet_model_architecture}).

\subsection{Key-to-Door: Long-Term Credit Assignment}
\label{app:key-to-door}

\paragraph{Model and Training Details:}

Model specifications and hyperparameters are outlined in Fig.~\ref{app:key_door_model_architecture}. CAT and baseline models were trained on 10,000 random walk trajectories and evaluated on 1,000 random starts. Baseline models included random policy, behavioral cloning, percentile behavior cloning (trained on winning trajectories only), conservative Q-learning, and decision transformers.

\begin{table}[t]
\caption{Model configurations for the Key-to-Door task.}
\label{app:key_door_model_architecture}
\centering
\small
\setlength{\tabcolsep}{2pt}
\resizebox{\linewidth}{!}{%
\begin{tabular}{lcccccccc}
\toprule
\textbf{Model} & \textbf{Variant} & \textbf{Params} & \textbf{Layers} & \textbf{Dim} & \textbf{Heads} & \textbf{MLP Dim} & \textbf{Context} & \textbf{LR} \\
\midrule
Random Policy & Base & 3M & 8 & 128 & 8 & 512 & 114 & 3e-3 \\
Behavioral Cloning & Base & 3M & 8 & 128 & 8 & 512 & 114 & 3e-3 \\
Percentile Behavioral Cloning & Base & 3M & 8 & 128 & 8 & 512 & 114 & 3e-3 \\
Conservative Q-Learning & Base & 3M & 8 & 128 & 8 & 512 & 114 & 3e-3 \\
Decision Transformers & Base & 3M & 8 & 128 & 8 & 512 & 114 & 3e-3 \\
CAT & Base & 3M & 8 & 128 & 8 & 512 & 114 & 3e-3 \\
\bottomrule
\end{tabular}%
}
\end{table}

\subsection{Language Modeling: Amazon Reviews}
\label{app:amazon-reviews}

\paragraph{Data Set:}

The Amazon Reviews data set (CC0: Public Domain) is a large-scale corpus of 574 million product reviews, each consisting of a product category, title, review text, and a 1–5 star rating \cite{hou2024bridging}. Reviews were tokenized using a 32k BPE vocabulary and formatted as \texttt{<|sos|><|category|><|sotitle|>title text<|sotext|>review text<|sor|><|*\_R|>}, where \texttt{R} denotes the rating. The data set was split 90:10 into training and validation sets. The distribution of ratings for 1–5 stars in the training set was 10.3\%, 4.9\%, 7.1\%, 12.7\%, and 65.0\%, respectively, with a similar distribution observed in the validation set.

\paragraph{Model and Training Details:}

We trained models ranging from 7-million to 1-billion parameters (Table~\ref{app:language_model_architecture}). For each model size, one standard decoder-only baseline and seven CAT variants with different values of $\lambda$ in Eq.~\ref{lossfn} were trained. CAT was also extended to a pre-trained standard model by fine-tuning only the attribute head. For DExperts, a 1-billion parameter standard model was used as the base model and a $\lambda = 0.15$ was used, as it outperformed the previously reported optimum of 0.2 on this dataset~\cite{arora2022director}.

\begin{table}[t]
\caption{Model configurations for the language modeling task.}
\label{app:language_model_architecture}
\centering
\small
\setlength{\tabcolsep}{2pt}
\begin{tabular}{lcccccccc}
\toprule
\textbf{Model} & \textbf{Variant} & \textbf{Params} & \textbf{Layers} & \textbf{Dim} & \textbf{Heads} & \textbf{MLP Dim} & \textbf{Context} & \textbf{LR} \\
\midrule
GPT & Base & 7M & 2 & 96 & 2 & 384 & 512 & 3e-4\\
GPT & Base & 72M & 12 & 512 & 8 & 2048 & 512 & 3e-4\\
GPT & Base & 270M & 16 & 1024 & 16 & 4096 & 512 & 3e-4\\
GPT & Base & 1B & 32 & 1536 & 32 & 6144 & 512 & 2.5e-4\\
CTRL & Base & 1B & 32 & 1536 & 32 & 6144 & 512 & 2.5e-4\\
DExperts & 1 star & 1B & 32 & 1536 & 32 & 6144 & 512 & 2.5e-4\\
DExperts & 5 star & 1B & 32 & 1536 & 32 & 6144 & 512 & 2.5e-4\\
Director & Binary (1 star) & 1B & 32 & 1536 & 32 & 6144 & 512 & 2.5e-4\\
Director & Binary (5 star) & 1B & 32 & 1536 & 32 & 6144 & 512 & 2.5e-4\\
Director* & Base & 1B & 32 & 1536 & 32 & 6144 & 512 & 2.5e-4\\
CAT & Base & 7M & 2 & 96 & 2 & 384 & 512 & 3e-4\\
CAT & Base & 72M & 12 & 512 & 8 & 2048 & 512 & 3e-4\\
CAT & Base & 270M & 16 & 1024 & 16 & 4096 & 512 & 3e-4\\
CAT & Base & 1B & 32 & 1536 & 32 & 6144 & 512 & 2.5e-4\\
CAT & Fine-tuned & 1B & 32 & 1536 & 32 & 6144 & 512 & 5e-2\\
CAT & Attribute-only & 1B & 32 & 1536 & 32 & 6144 & 512 & 5e-2 \\
\bottomrule
\\
\end{tabular}
\end{table}

\paragraph{Evaluation Details:}

\begin{figure}[htbp]
    \centering
    \includegraphics[width=1.0\linewidth]{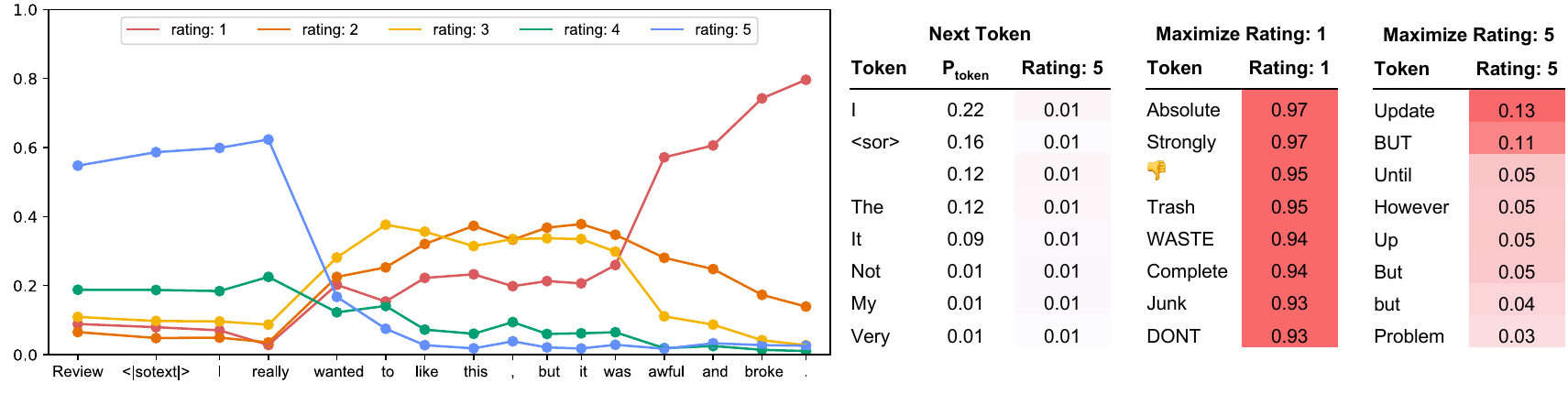}
    \caption{Token-level attribute estimates. CAT serves as a token-level critic of reviews, estimating attributes using the conditional attribute probabilities for the true next token. The model also estimates counterfactual probabilities that can be used to maximize the probability of any star review (1 star and 5 star shown).}
    \label{app:reviewoverview}
\end{figure}

\begin{figure}[htpb]
    \centering
    \includegraphics[width=1.0\linewidth]{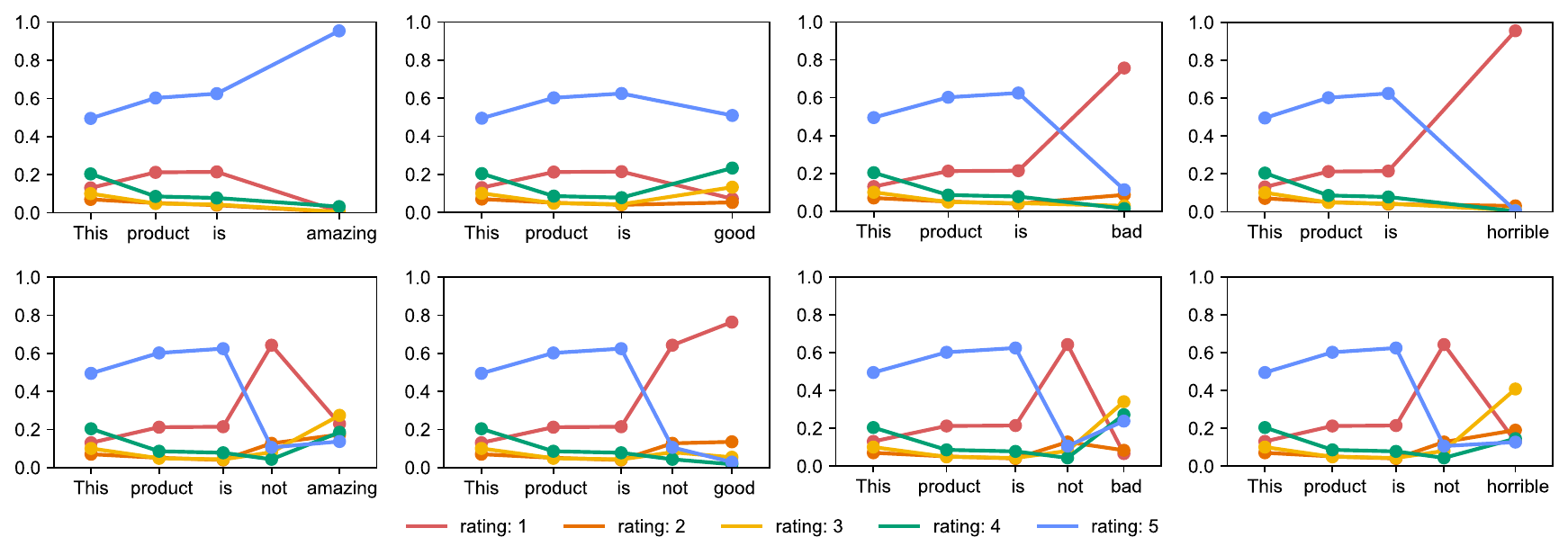}
    \caption{Token-level attribute estimates for adjective substitution. Token-level attribute estimates from CAT demonstrating its ability to capture sentiment and the contextual effect of negation across the adjective substitutions \textit{amazing}, \textit{good}, \textit{bad}, and \textit{horrible}.}
    \label{app:reviewdemonstration}
\end{figure}

Next-token validation perplexity was evaluated over 2,000 iterations for the standard and CAT models across model sizes. 

Counterfactual adjective substitution was evaluated on 1,000,000 validation reviews in which the true next token was \textit{good}. Results are reported for all reviews and for the subset of 1,593 reviews with a preceding negation term or phrase: \textit{not}, \textit{no}, \textit{never}, \textit{not really}, \textit{isn’t}, \textit{ain’t}, \textit{wasn’t}, \textit{weren’t}. Fig.~\ref{app:reviewdemonstration} visualizes the conditional attribute estimates at each token position. \textit{This product is amazing}, \textit{This product is good}, \textit{This product is bad}, and \textit{This product is horrible} produce similar probability distributions up to the adjective token, after which the predicted rating probabilities diverge according to adjective sentiment. Positive adjectives increase the predicted probability of 5 star ratings, whereas negative adjectives increase the predicted probability of 1 star ratings. Stronger adjectives (\textit{amazing}, \textit{horrible}) produce larger shifts than weaker adjectives (\textit{good}, \textit{bad}). A similar pattern is observed under negation. The reviews produce similar distributions up to \textit{not}, after which the predicted probability of a 5 star rating decreases and that of a 1 star rating increases, indicating that the model has learned to associate negation with more negative sentiment and lower ratings. Adjectives following negation produce effects consistent with those of negation shown in Table~\ref{tab:counterfactual_estimation}. \textit{This product is not good} resembles \textit{This product is bad} with an increased predicted probability of a 1 star rating, while \textit{not bad}, \textit{not amazing}, and \textit{not horrible} are dominated by a 3 star rating, consistent with indicating \textit{okay}.

\begin{figure}[htpb]
    \centering
    \includegraphics[width=0.5\linewidth]{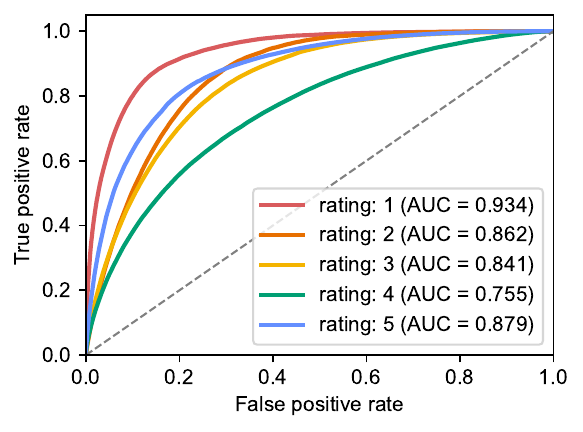}
    \caption{AUC-ROC curve for XGBoost critic model across all five ratings.}
    \label{fig:xgboost}
\end{figure}

For the steering experiment, validation reviews with true 3 star ratings were split after the first 50\% of the text following the \texttt{<|sotext|>} token. The resulting first-half prompts were scored using an XGBoost rating classifier trained on 10,000,000 validation reviews with CountVectorizer features (Fig.~\ref{fig:xgboost}) \cite{chen2026xgboost}. Prompts classified as 3 stars were retained, from which 1,000 were randomly selected. Steering was performed using a satisficing criterion with top-$k$ decoding for CAT, Director, and Director* models. At each step, decoding was restricted to tokens with a next-token probability above $\epsilon$ and a conditional attribute probability above the specified attribute threshold~\cite{simon1955behavioral}. DExperts scales the difference between expert and anti-expert logits with a control parameter $\alpha$; following prior work, we use $\alpha=0.2$ for steering~\cite{liu2021dexperts}. Hyperparameters for all model types are listed in Table~\ref{app:steering_hyperparams}. Each method generated 10 completions per prompt, and the resulting full text, consisting of the prompt and completion, was re-scored by XGBoost. Performance is reported as the proportion of full texts classified as either 1 or 5 stars. Fluency of the completion was measured by mean perplexity using a Hugging Face GPT-2 XL model and tokenizer~\cite{radford2019language}, and diversity of the completion was measured using normalized unique $n$-gram counts. Evaluation metrics closely follow those used for the toxicity task in~\cite{liu2021dexperts}.

\begin{table}[t]
\caption{Steering hyperparameters.}
\label{app:steering_hyperparams}
\centering
\small
\setlength{\tabcolsep}{6pt}
\begin{tabular}{lccccc}
\toprule
\textbf{Method} & $\boldsymbol{k}$ & $\boldsymbol{\alpha}$ & $\boldsymbol{\epsilon}$ & \textbf{\makecell{Attribute\\Threshold}} & \textbf{\makecell{Sampling\\Method}} \\
\midrule
CTRL       & 20 & --  & --    & --  & Top-$k$ \\
DExperts   & 20 & 0.2 & --    & --  & Top-$k$ \\
Director   & 20 & --  & 0.001 & 0.8 & Top-$k$ \\
Director*  & 20 & --  & 0.001 & 0.8 & Top-$k$ \\
CAT        & 20 & --  & 0.001 & 0.8 & Top-$k$ \\
\bottomrule
\end{tabular}
\end{table}

\subsection{PhysioNet Sepsis}

\paragraph{Data Set:}

The 2019 PhysioNet Challenge (CC-BY-4.0) provides a data set from 40,336 patients collected from the Intensive Care Units (ICUs) of three hospital systems as a benchmark for early sepsis identification \cite{reyna2019early}. This data set consists of patient demographic information and hourly measurements of vital signs and laboratory values. Each ICU course is annotated with whether sepsis occurred and its time of onset. Records were converted into sequences of discrete tokens. Patient demographics were included at the beginning of each sequence. Categorical measurements received a unique discrete token and each numeric measurement received binned value tokens. Values were discretized into deciles based on per-class distribution of values in the training set. A time token was inserted between each set of hourly measurements. Data was split 90:10 into training and validation sets, with a sepsis prevalence of 7.3\% in training and 7.4\% in validation. Model hyperparameters are listed in Table~\ref{app:physionet_model_architecture}.

\paragraph{Model and Training Details:}

The optimal $\lambda$ used to balance the contribution of next-token and target attribute loss differed in the binary attribute prediction and regression target in the sepsis data set. We selected a $\lambda$ of 0.01 for binary sepsis risk prediction and 0.5 for predicting the regression target of the max heart rate in the next 6 hours. The large difference in these values highlights how predicting the next token correctly is essential for sepsis risk prediction, whereas a larger contribution of target regression loss is necessary to learn the more volatile value of heart rate.

\begin{table}[t]
\caption{Model configurations for the sepsis prediction task.}
\label{app:physionet_model_architecture}
\centering
\small
\setlength{\tabcolsep}{2pt}
\begin{tabular}{lcccccccc}
\toprule
\textbf{Model} & \textbf{Variant} & \textbf{Params} & \textbf{Layers} & \textbf{Dim} & \textbf{Heads} & \textbf{MLP Dim} & \textbf{Context} & \textbf{LR} \\
\midrule
GPT & Base & 29M & 8 & 512 & 8 & 2048 & 1024 & 5e-4\\
Director & Base & 29M & 8 & 512 & 8 & 2048 & 1024 & 5e-4\\
CAT & Binary & 29M & 8 & 512 & 8 & 2048 & 1024 & 5e-4\\
CAT & Regression & 29M & 8 & 512 & 8 & 2048 & 1024 & 5e-5\\
\bottomrule
\\
\end{tabular}
\end{table}

\begin{figure}[htpb]
    \centering
    \includegraphics[width=1.0\linewidth]{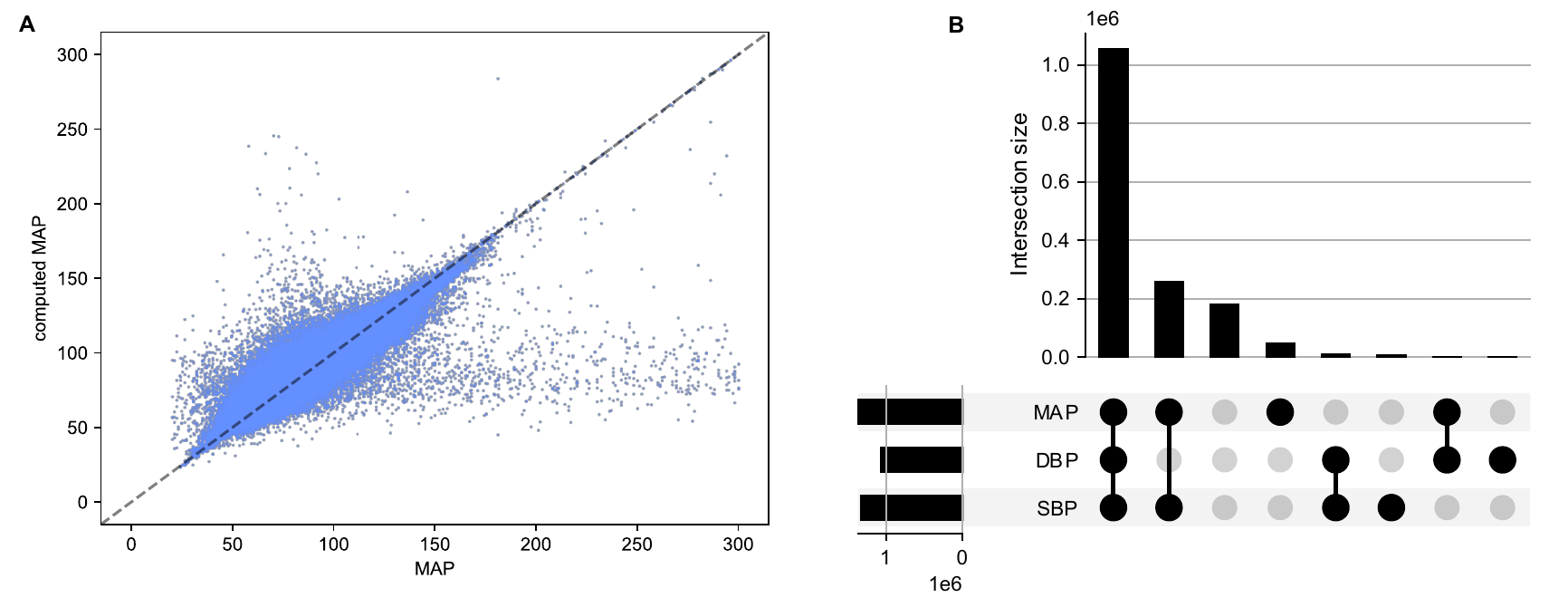}
    \caption{Blood Pressure reading and correlation to reported MAP. \textbf{A} Scatter plot comparing computed MAP to reported MAP. \textbf{B} UpSet plot highlighting overlap of hours where SBP, DBP, and MAP are reported.}
    \label{fig:sepsissupplement}
\end{figure}


\end{document}